\documentclass[twocolumn, switch]{article} % Method A for two-column formatting

\usepackage{preprint}

\usepackage{amsmath, amsthm, amssymb, amsfonts}

\usepackage[numbers,square]{natbib}
\usepackage[utf8]{inputenc}	% allow utf-8 input
\usepackage[T1]{fontenc}	% use 8-bit T1 fonts
\usepackage{xcolor}		% colors for hyperlinks
\usepackage{hyperref}	% Color links to references, figures, etc.
\usepackage{booktabs} 		% professional-quality tables
\usepackage{nicefrac}		% compact symbols for 1/2, etc.
\usepackage{microtype}		% microtypography
\usepackage{lineno}		% Line numbers
\usepackage{float}			% Allows for figures within multicol

\usepackage{lipsum}		%  Filler text

\usepackage{newfloat}
\DeclareFloatingEnvironment[name={Supplementary Figure}]{suppfigure}
\usepackage{sidecap}
\sidecaptionvpos{figure}{c}

\usepackage{titlesec}
\titlespacing\section{0pt}{12pt plus 3pt minus 3pt}{1pt plus 1pt minus 1pt}
\titlespacing\subsection{0pt}{10pt plus 3pt minus 3pt}{1pt plus 1pt minus 1pt}
\titlespacing\subsubsection{0pt}{8pt plus 3pt minus 3pt}{1pt plus 1pt minus 1pt}

\title{Modification method for single-stage object detectors that allows to exploit the temporal behaviour of a scene to improve detection accuracy}

\usepackage{xwatermark}
\newwatermark[firstpage,color=gray!90,angle=0,scale=0.28, xpos=0in,ypos=-5in]{*correspondence: \texttt{gmenua@gmail.com}}

\usepackage{authblk}

\author[1]{Menua Gevorgyan}
\affil[1]{ Institute for Physical Research of National Academy of Sciences of Armenia}
\begin{document}

\twocolumn[ % Method A for two-column formatting
  \begin{@twocolumnfalse} % Method A for two-column formatting

\maketitle

\begin{abstract}

A simple modification method for single-stage generic object detection neural networks, such as YOLO and SSD, is proposed, which allows for improving the detection accuracy on video data by exploiting the temporal behavior of the scene in the detection pipeline. It is shown that, using this method, the detection accuracy of the base network can be considerably improved, especially for occluded and hidden objects. It is shown that a modified network is more prone to detect hidden objects with more confidence than an unmodified one. A weakly supervised training method is proposed, which allows for training a modified network without requiring any additional annotated data.

\end{abstract}

\vspace{0.35cm}

  \end{@twocolumnfalse} % Method A for two-column formatting
] % Method A for two-column formatting

\section*{Introduction}

Recently, multiple convolutional neural network based generic object detection architectures arose, which allowed to dramatically improve the detection accuracy relative to traditional approaches and achieve state-of-the-art performance. These neural architectures allowed for creating multiple practical systems in the field of robotics, automation, and monitoring. These architectures can mainly be categorised into two types, i.e., single- and two-stage detectors. The latter are region proposal neural networks and include the R-CNN family \citep{rcnn,fast-rcnn,faster-rcnn,mask-rcnn}, SPP-net \citep{spp-net}, R-FCN \citep{r-fcn}, and FPN \citep{fpn}. Meanwhile, single-stage object detectors predict all objects in a scene with a single pass through the neural network.
Such dectors include Yolo \citep{yolov3}, SSD \citep{ssd}, and SqueezeDet \citep{squeezedet}.

All these neural architectures have multiple practical applications, particularly, they are often used for training and making inferences on still frames extracted from video streams. Meanwhile, a priori, they can only detect objects in a video frame that can be detected by looking at that video frame. Although, in many scenarios, the information available in the current frame is not enough for reliably detecting an object. For example, an object, which is occluded in the current frame, cannot be detected by looking only into that current frame; however, if we also look at the previous frames, where the object was not occluded, we can be confident about the location of the object in the current frame.

In this work, we propose a new modification method for single-stage object detectors.
This method aims to encode the previous frames of a video in a way that can be exploited to improve detections in the current frame.

We show that this modification method allows for a considerable improvement in the object detection accuracy of the base model, especially, for occluded objects. We also show that a modified network is more prone to produce higher scores for occluded objects than the base network.

In a related work \citep{mobile_lstm},  a recurrent convolutional architecture was created by replacing convolutional layers of the base single-stage object detection network with the proposed Bottleneck-LSTM layers. These  Bottleneck-LSTM are similar to  ConvLSTM  layers \citep{convlstm} and use depthwise separable convolution instead of the simple one, which makes it possible to reduce computational costs dramatically. In contrast to the latter work, in our work, we use ConvLSTM cells to encode information from previous frames only, without including the current frame. Meanwhile, the features of the unmodified architecture, representing the current frame, are used together with recurrent encoding of the previous frames in further computations. The motivation behind our approach is that, in our case, the recurrent unit is not forced to encode the current frame, which we suppose on average contains relatively more information for making reliable predictions than the information found in the previous frames. The latter allows for using the recurrent unit more efficiently for encoding frame history. Another motivation is that our approach makes it possible to transfer all weights from the original architecture. The work \citep{tssd} proposes an architecture that is based on an attention mechanism and ConvLSTM units to accomplish temporally aware single-shot real-world object detection.

In this work, we also propose a weakly supervised training approach for a modified network. This approach allows for training the model on a dataset of videos, which only provides annotations for separate isolated frames of a video sequence as apposed to annotations for all frames of a dataset. This training approach permits to use the modification method in practical systems that apply single-stage object detectors and are trained on frames taken from videos, without requiring additional data.

\section{Modification Method}

\begin{figure}[t]
\centering
\includegraphics[width=0.9\columnwidth]{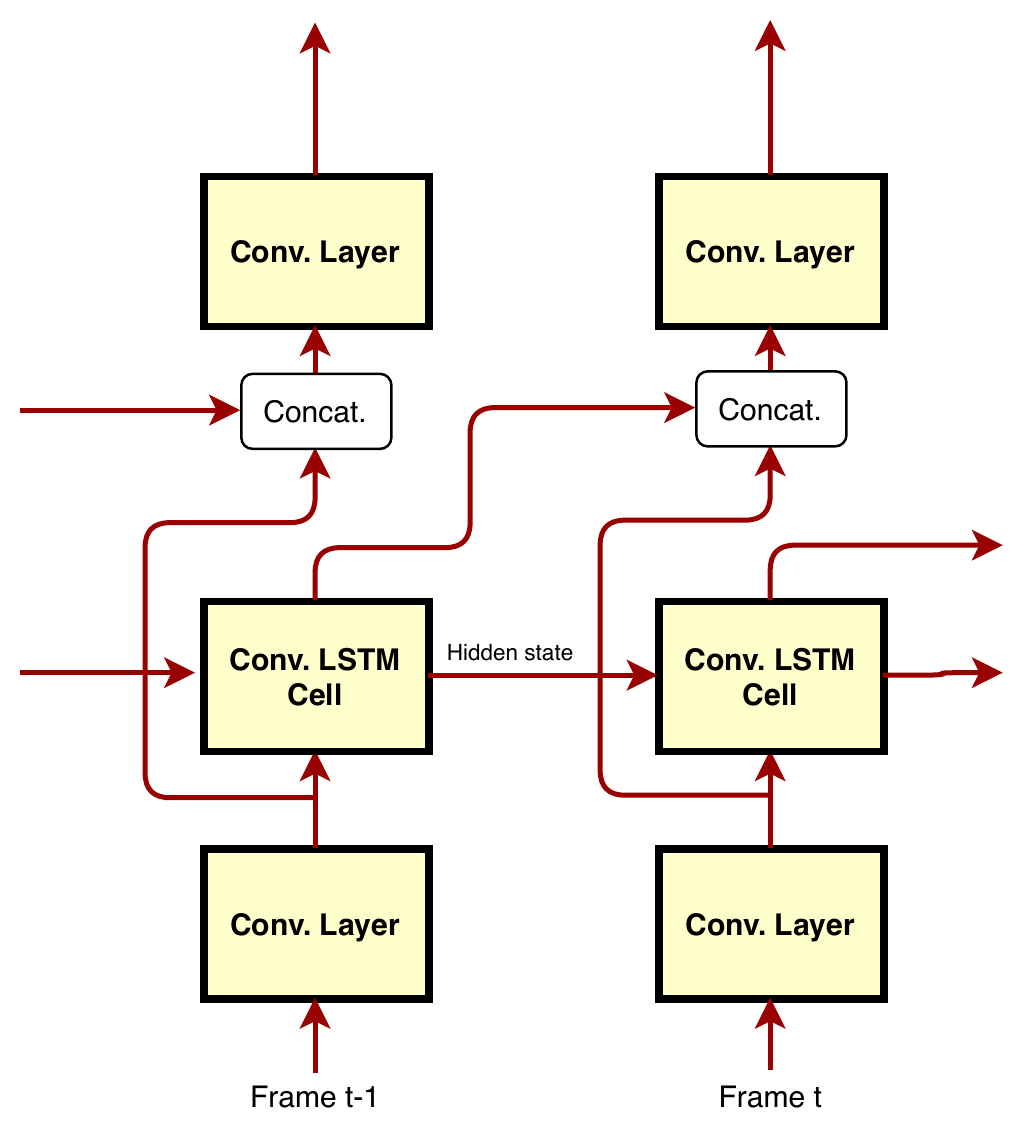} % Reduce the figure size so that it is slightly narrower than the column. Don't use precise values for figure width.This setup will avoid overfull boxes.
\caption{Modification is done between two convolutional layers. A ConvLSTM cell encodes output feature maps of the lower convolutional layer up to, but not including, feature maps for the current frame. This encoding is concatenated with the output of the lower layer computed on the current frame and passed to the upper convolutional layer for further processing.}
\label{f:nn_modification}
\end{figure}

We propose a modification method for single-stage object detectors, such as Yolo \citep{yolov3}, SSD \citep{ssd}, and SqueezeDet \citep{squeezedet}, which aims to embed the history in such a way that can be exploited to improve detections for a current frame. This mechanism embeds features of certain convolutional layers of a neural network and their time behaviour in the previous frames, using a recurrent architecture, namely, ConvLSTM \citep{convlstm} cells. Next, this encoding is passed to the upper layer along with feature maps of the same layer of the unmodified network, as illustrated in the figure \ref{f:nn_modification}.

\section{Experimental Setup}

In our experimental setup, we use YOLOv3-tiny \citep{yolov3} as a base network, as it is lightweight, quick to train, and, at the same time, very robust.
We use a TensorFlow 2.0 implemetation of the YOLO3-tiny network available on GitHub\footnote{YoloV3 Implemented in TensorFlow 2.0: \url{https://github.com/zzh8829/yolov3-tf2}.}.
We achieve this modification by inserting ConvLSTM \citep{convlstm} cells at the last feature maps of the neural network, just before the prediction output layers. YOLOv3-tiny has two such outputs. Each of these outputs is modified with a separate ConvLSTM cell with a number of filters equal to the number of filters of the encoded layer. Next, the output of the ConvLSTM cell is concatenated with the output of the same layer of the modified network. As a result, the number of the input feature maps of the output layer doubles. Such an experimental setup only slightly increases the computational costs required for forward propagation through the network.

\section{Training}

We train our neural network on the MOT16-17 dataset \citep{mot}. This is a convenient dataset for evaluating the performance of a network on occluded objects, as it contains corresponding ground truth. We use the first 80\% of each video in the dataset for training, while the remaining 20\% is for testing.
The original network is pretrained on ImageNet \citep{imagenet} and COCO \citep{coco} datasets.
The input to the neural network is a sequence of video frames, while the supervisory signal is composed of an annotation on a single frame of the sequence. The location of that frame in the sequence is chosen randomly. This training approach allows to train a model on a video dataset, which only provides ground-truth bounding boxes for isolated frames from video sequences instead of all frames in a sequence. We train using the same loss function as for an unmodified network.

In our training process, we freeze all layers preceding the convolutional layers, after which we introduce the modification.
When training large recurrent neural networks, one of the main issues is related to fitting network states through time into the memory of the computing unit in order to compute the gradient of all weights of a neural network.
As a result of the freezing of most layers of the base network, it is only required to keep the computational results of the nonfrozen layers, while backpropagating through time. Such a reduction in the volume of required information makes it possible to select large neural architectures as a base network, such as YoloV3, and use a larger number of consequent frames during training without exceeding the memory limits of the computing unit.

We train the network for 10 epochs without changing hyperparameters in the implemetation.

\section{Evaluation Metrics}

During evaluation, we aim to compare the relative performance of the model when it encodes information from previous frames with a situation when it only uses the current frame to make predictions. We introduce a notion that the model predicts in plain and sequenced modes, when it respectively uses a single frame or also encodes the previous ones to make predictions.
For evaluation purposes, we compare all-point interpolated Precision x Recall curves and all-point interpolated average precisions (AP) \citep{metric_implementation}. All evaluations are done for $70\%$ of intersection over the uninon threshold.

Precision $P$ and recall $R$ are  defined as follows:
\begin{equation}
    \label{eq:precision}
    P = \frac{TP}{TP+FP},
\end{equation}

\begin{equation}
    \label{eq:recall}
    R = \frac{TP}{TP+FN},
\end{equation}

where $TP$ is the true positive (the number of correct detections of ground-truth
bounding boxes); $FP$ is the false negative (the number of incorrect detections of bounding boxes); and $FN$ is the false negative (the number of undetected ground-truth bounding boxes).

We are also interested in comparing the performances between plain and sequenced inference modes on objects which are occluded by other objects in the scene and those that are visible. For this purpose, we utilize the visibility ratio provided in the ground truths of the MOT17 dataset, which is a number between 0 and 1 that indicates which part of an object is visible. This number varies due to occlusions and image border cropping.

Using this information, we split true positives on two parts:
\begin{equation}
    \label{simple_equation}
    TP = TP_{hidden} + TP_{visible},
\end{equation}
where $TP_{hidden}$ is the number of true positive bounding boxes with visibility less than $0.5$ and $TP_{visible}$ is the number of true positive bounding boxes with visibility greater than or equal to $0.5$. Next, we define the corresponding hidden and visible precisions and recalls as

\begin{equation}
    \label{eq:pr_visible}
    P_{visible} = \frac{TP_{visible}}{TP_{visible}+FP},
    R_{visible} = \frac{TP_{visible}}{TP_{visible}+FN}
\end{equation}

\begin{equation}
    \label{eq:pr_hidden}
    P_{hidden} = \frac{TP_{hidden}}{TP_{hidden}+FP},
    R_{visible} = \frac{TP_{hidden}}{TP_{hidden}+FN}
\end{equation}

and use them to calculate the corresponding Precision x Recall curves and average precisions.

\section{Results}

% \begin{figure}[h]
% \begin{center}
% %\framebox[4.0in]{$\;$}
% \fbox{\rule[-.5cm]{0cm}{4cm} \rule[-.5cm]{4cm}{0cm}}
% \end{center}
% \caption{Sample figure caption.}
% \end{figure}
\newcommand{\fprc}{All-point interpolated Precision x Recall curves for plain and sequend modes }
\begin{figure}[t]
\centering
\includegraphics[width=0.9\columnwidth]{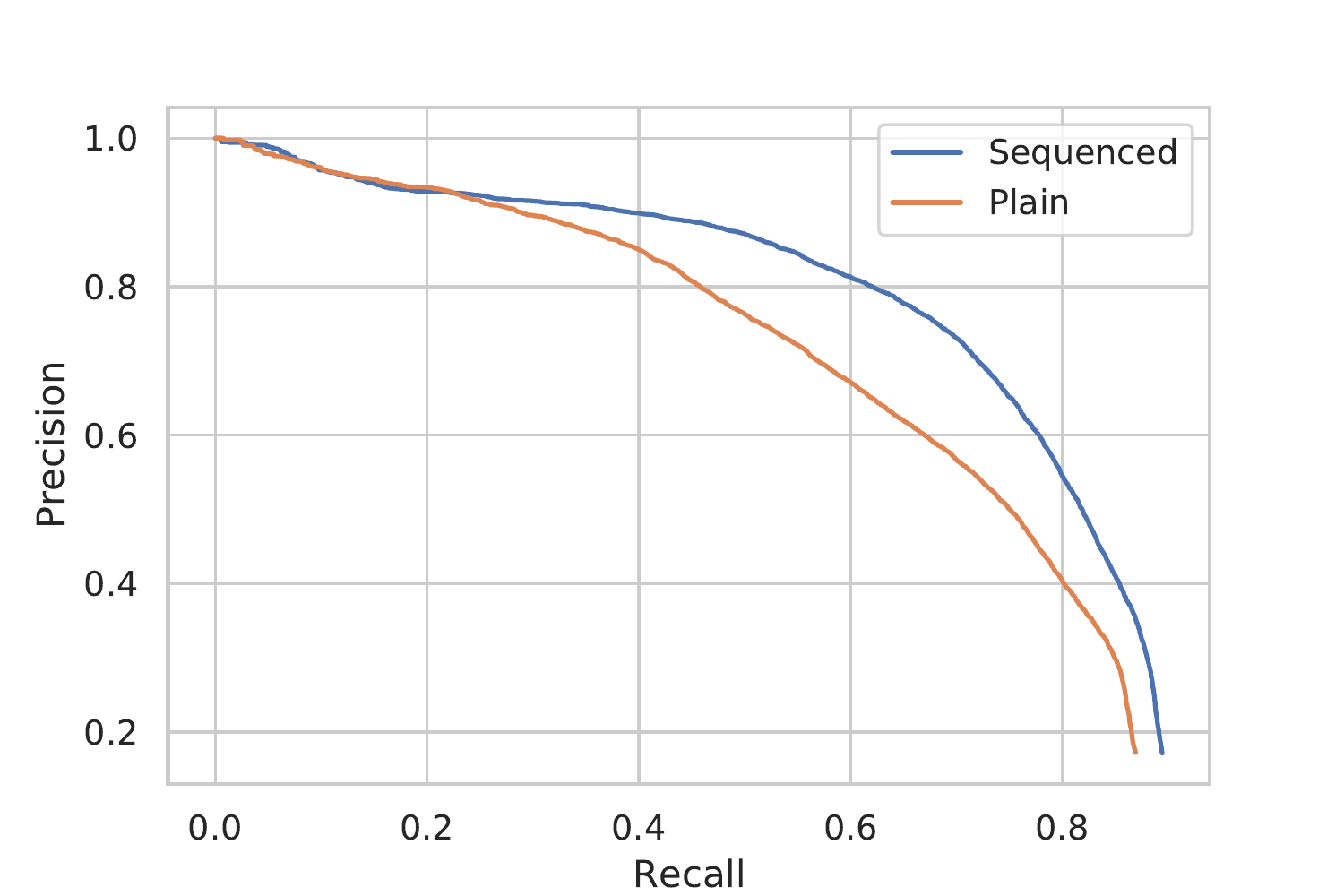} % Reduce the figure size so that it is slightly narrower than the column. Don't use precise values for figure width.This setup will avoid overfull boxes.
\caption{\fprc.}
\label{f:all}
\end{figure}

\newcommand{\prc}{interpolated Precision x Recall curves }

Figure \ref{f:all} demonstrates \prc on the test dataset when the model runs in plain and sequenced modes. The model has similar precisions until recall $~0.2$ in plain and sequenced modes. Meanwhile, above recall $~0.2$, in the case when the model predicts in the sequenced mode, it considerably outperforms its predictions in the plain mode. The AP values for the plain and sequenced modes are $0.66$ and $0.73$, respectively, which indicates that the sequenced mode outperforms the plain mode by approximately $20\%$ in terms of the absolute error according to the AP metric.

\begin{figure}[t]
\centering
\includegraphics[width=0.9\columnwidth]{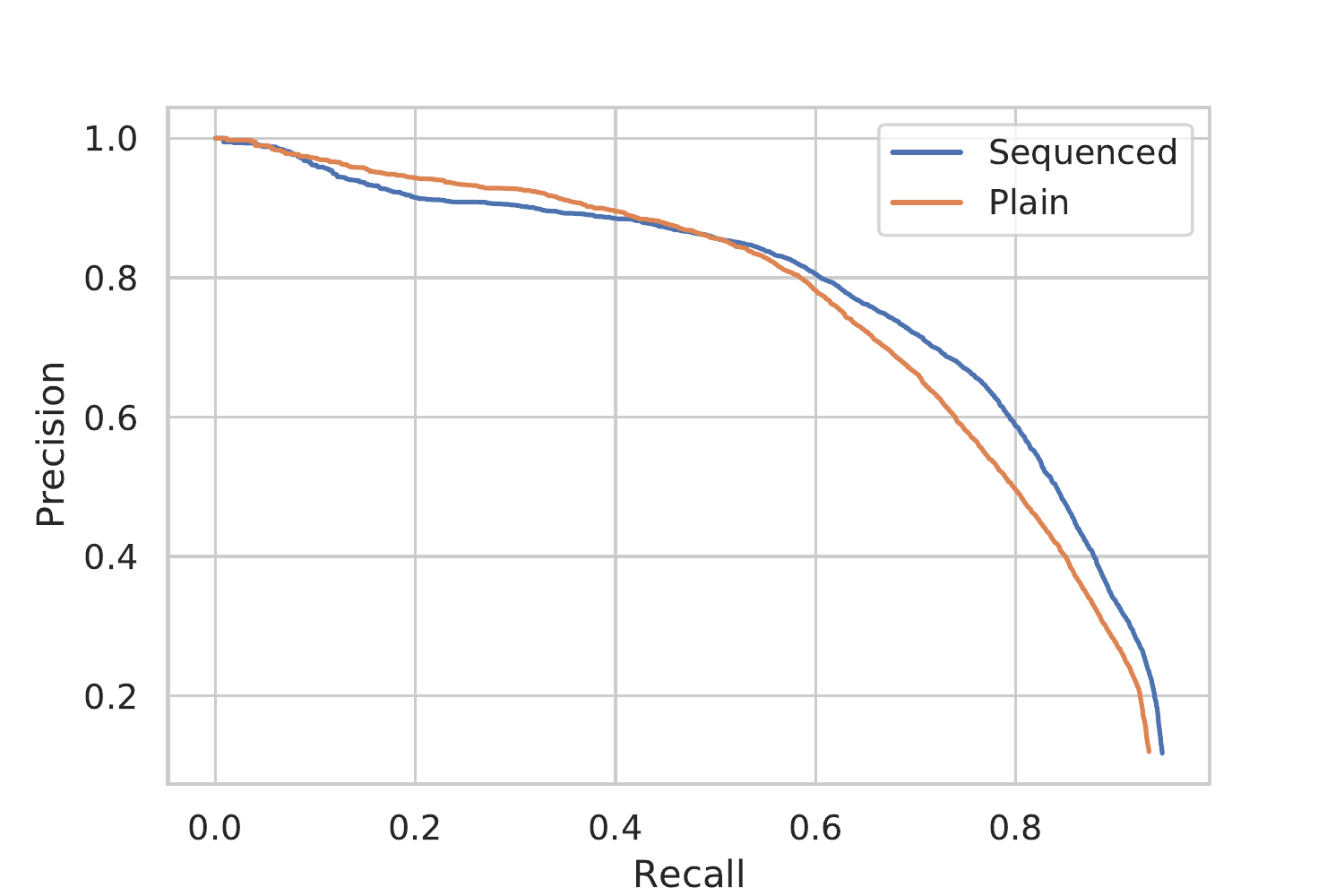} % Reduce the figure size so that it is slightly narrower than the column. Don't use precise values for figure width.This setup will avoid overfull boxes.
\caption{\fprc for visible objects (according to definitions (\ref{eq:pr_visible})).}
\label{f:shown}
\end{figure}

Figure \ref{f:shown} shows \prc for the visible objects in the test dataset according to precision and recall definitions (\ref{eq:pr_visible}). Overall, both inference modes have similar performance on visible objects. It can be noted that, according to the AP score and definition (\ref{eq:pr_visible}), the plain mode outperforms slightly the sequenced mode on visible objects for the recall $<0.5$, and, vice versa, the sequenced mode outperforms slightly the plain mode for the recall $>0.5$. The AP values for plain and sequenced modes are $0.72$ and $0.74$, respectively, showing that, on average, the sequenced inference mode outperforms slightly the plain one.

\begin{figure}[t]
\centering
\includegraphics[width=0.9\columnwidth]{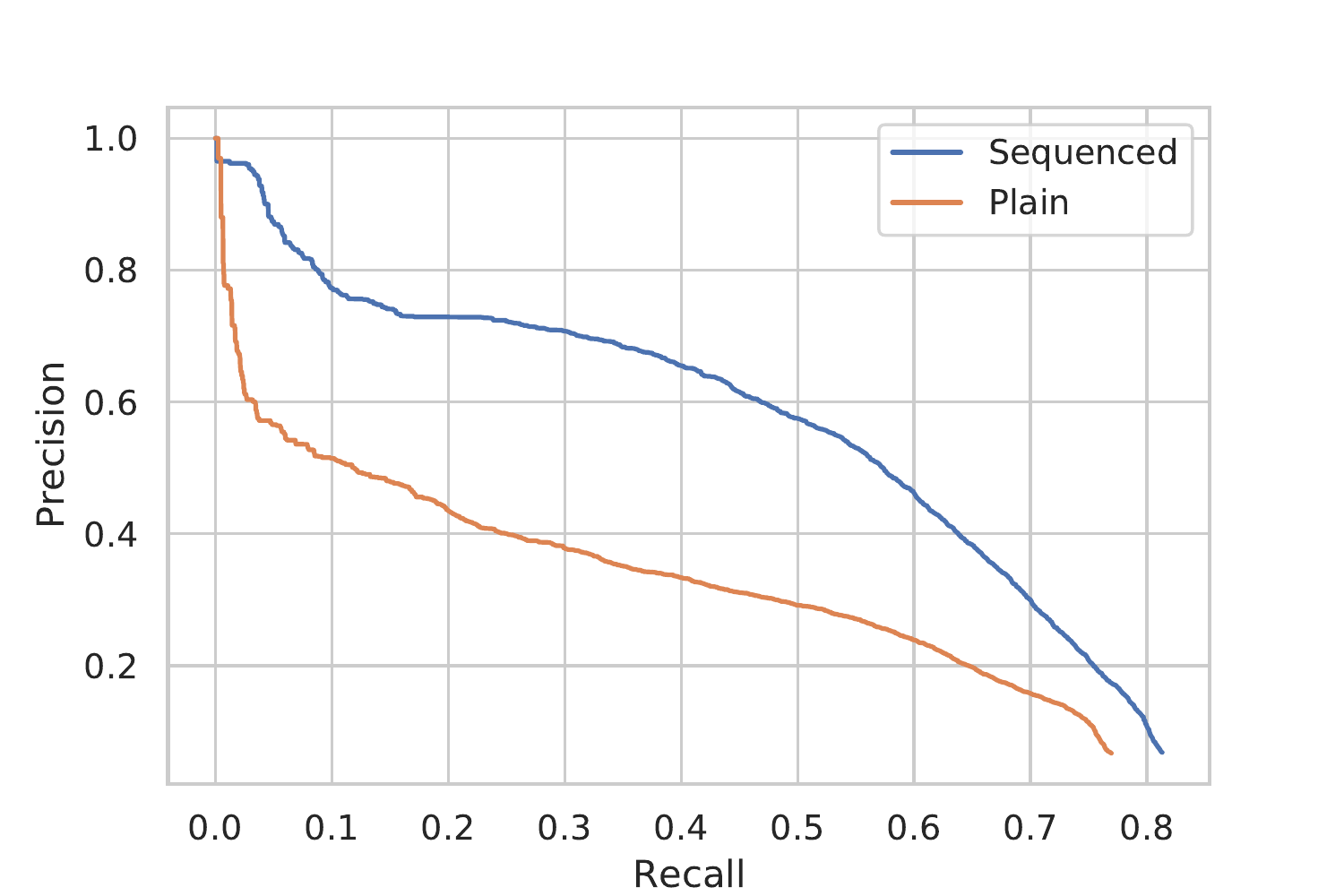} % Reduce the figure size so that it is slightly narrower than the column. Don't use precise values for figure width.This setup will avoid overfull boxes.
\caption{\fprc for hidden objects (according to definitons (\ref{eq:pr_hidden})).}
\label{f:hidden}
\end{figure}

Figure (\ref{f:hidden}) demonstrates \prc for the hidden objects in the test dataset according to precision and recall definitions (\ref{eq:pr_hidden}). We can note that the sequenced inference outperforms enormously the plain inference mode for hidden objects, which is in contrast to the case of visible objects. The average precisions are $0.27$ and $0.48$, respectively, which indicates that in the sequenced inference mode the network is able to detect correctly almost twice as many hidden objects as in the plain inference mode. The latter shows that the network can encode information from previous frames in the video stream in such a way that it is pretty confident about the locations of the objects that become hidden.

\begin{figure}[t]
\centering
\includegraphics[width=0.9\columnwidth]{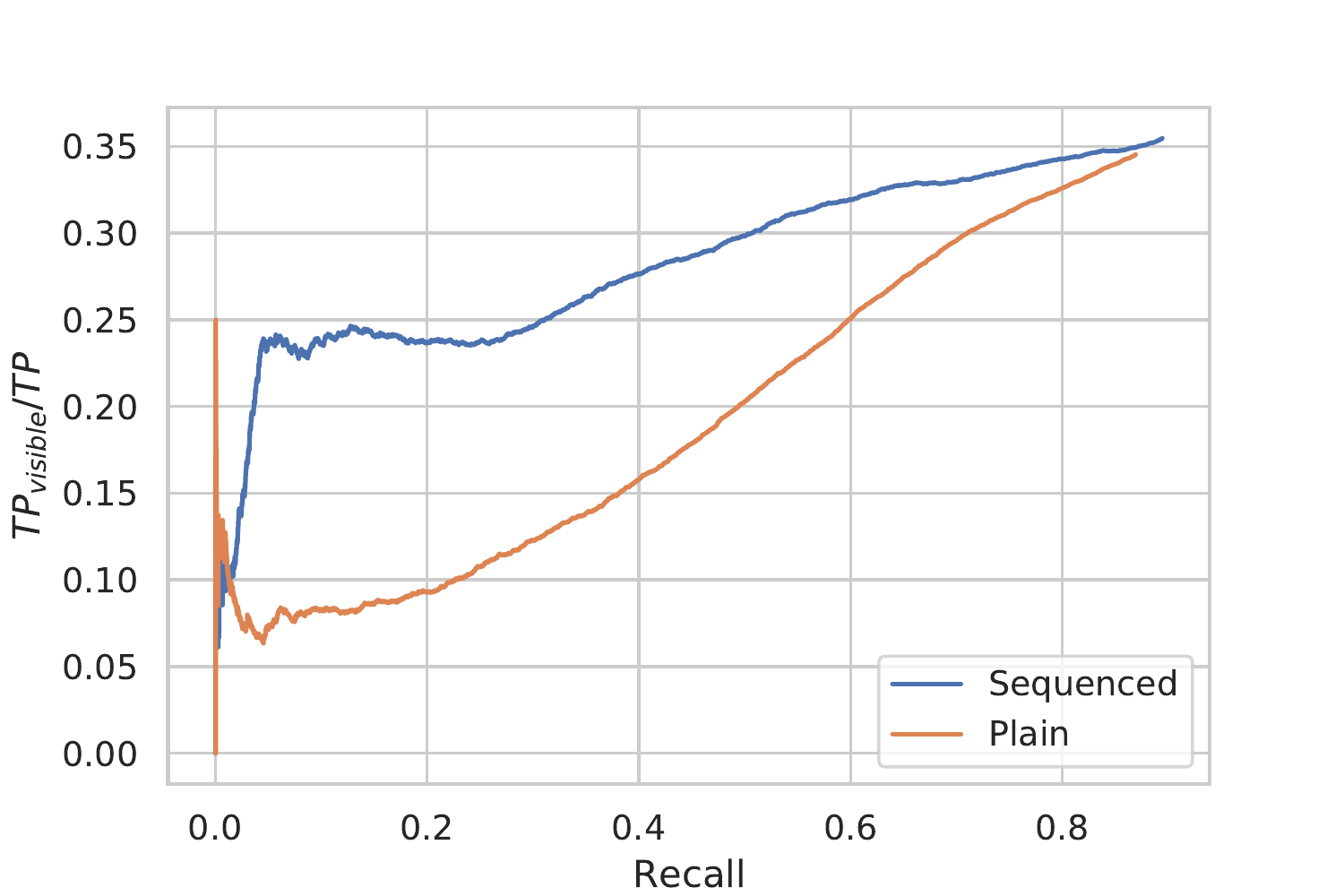} % Reduce the figure size so that it is slightly narrower than the column. Don't use precise values for figure width.This setup will avoid overfull boxes.
\caption{$TP_{visible}/TP$ vs Recall for plain and sequenced inference modes.}
\label{f:prone}
\end{figure}

Figure \ref{f:prone} shows the $TP_{visible}/TP$ value relative to the recall for the plain and sequenced inference modes. The curves demonstrate how prone the network is to produce a high probability to hidden objects.
 % relatve to visible ones in plain and sequenced inference mode respectively.
 It can be noted that,
 after recalling a small number of ground truth bounding boxes,
 the network gives considerably higher scores to hidden objects when it runs in the sequenced mode as compared to its runs in the plain mode. Both curves rise with increasing recall values. The relation between the total numbers of hidden and all ground truth boxes is equal to $0.39$, which is the point where both curves would reach $R=1$ if the network was able to recall all ground truth boxes.

\section{Conclusion}
% Write your conclusion here.
% mainstream?

In this paper, we have proposed a modification method for single-stage generic object detection convolutional neural networks. This method allows to encode the temporal behaviour of a scene, which serves as additional information in the object detection pipeline. The application of this method allows for considerable improvements in the accuracy of the base network, while it only slightly increases computational costs.
We have compared the inference on still frames with that on a video stream and have shown dramatic improvements in occluded object detection, as well as a considerable overall detection improvement.

Nowadays, most state-of-the-art object detection architectures make predictions by looking at still frames and find multiple industrial applications by processing each frame of a video stream. We have also proposed a training approach, which makes it possible to train a modified network on existing data, without requiring tedious annotations of all frames of the video dataset. The simultaneous application of the modification method and the corresponding training approach
makes it  easy to improve products that are largely based on object detection.
Industrial fields of such products include, but are not limited to, security monitoring, smart traffic light systems, manufacturing automation, and medicine.

In contrast to standard object detection approaches, people tend to take into account time behavior when they point to the location of an object. For example, if we observe a person walk around a corner of a building, we can be pretty sure about their location, although they are out of sight at this stage. From this point of view, we can consider this work to be an attempt to model temporal considerations of human visual perception.

\bibliography{refs}

\end{document}